\documentclass[sigconf,screen]{acmart}
\acmSubmissionID{401}
\usepackage{mmstyle}
\usepackage{booktabs} 

\usepackage{enumitem}

\usepackage[ruled]{algorithm2e} 

\SetAlFnt{\small}
\SetAlCapFnt{\small}
\SetAlCapNameFnt{\small}
\SetAlCapHSkip{0pt}

\acmJournal{TOG}

\setcopyright{acmlicensed}

\usepackage{bbm}
\newcommand{\ourmodel}[1]{HyperDreamer}

\copyrightyear{2023}
\acmYear{2023}
\setcopyright{rightsretained}
\acmConference[SA Conference Papers '23]{SIGGRAPH Asia 2023 Conference Papers}{December 12--15, 2023}{Sydney, NSW, Australia}
\acmBooktitle{SIGGRAPH Asia 2023 Conference Papers (SA Conference Papers '23), December 12--15, 2023, Sydney, NSW, Australia}
\acmDOI{10.1145/3610548.3618168}
\acmISBN{979-8-4007-0315-7/23/12}

\begin{document}
\title{HyperDreamer: Hyper-Realistic 3D Content Generation and Editing from a Single Image}

\author{Tong Wu}
\orcid{0000-0001-5557-0623}
\authornote{Equal contribution.}
\email{wt020@ie.cuhk.edu.hk}
\affiliation{%
    \institution{The Chinese University of Hong Kong}
	\country{China}
}
\affiliation{%
    \institution{Shanghai AI Laboratory}
	\country{China}
}
\author{Zhibing Li}
\orcid{0009-0002-4528-5495}
\authornotemark[1]
\email{lz022@ie.cuhk.edu.hk}
\affiliation{%
    \institution{The Chinese University of Hong Kong}
	\country{China}
}
\affiliation{%
    \institution{Shanghai AI Laboratory}
	\country{China}
}
\author{Shuai Yang}
\orcid{0009-0008-9552-4320}
\authornotemark[1]
\email{yssss.mikey@gmail.com}
\affiliation{%
    \institution{Shanghai AI Laboratory}
	\country{China}
}
\affiliation{
    \institution{Shanghai Jiao Tong University}
    \country{China}
}
\author{Pan Zhang}
\orcid{0009-0004-7195-4159}
\email{zhangpan@pjlab.org.cn}
\affiliation{%
	\institution{Shanghai AI Laboratory}
	\country{China}
}
\author{Xingang Pan}
\orcid{0000-0002-5825-9467}
\email{xingang.pan@ntu.edu.sg}
\affiliation{%
	\institution{S-Lab, NTU}
	\country{Singapore}
}
\author{Jiaqi Wang}
\orcid{0000-0001-6877-5353}
\email{wangjiaqi@pjlab.org.cn}
\affiliation{%
	\institution{Shanghai AI Laboratory}
	\country{China}
}
\author{Dahua Lin}
\authornote{Corresponding Authors.}
\orcid{0000-0002-8865-7896}
\email{dhlin@ie.cuhk.edu.hk}
\affiliation{%
    \institution{The Chinese University of Hong Kong}
	\country{China}
}
\affiliation{%
    \institution{Shanghai AI Laboratory}
	\country{China}
}
\author{Ziwei Liu}
\authornotemark[2]
\orcid{0000-0002-4220-5958}
\email{ziwei.liu@ntu.edu.sg}
\affiliation{%
	\institution{S-Lab, NTU}
	\country{Singapore}
}

\begin{abstract}
3D content creation from a single image is a long-standing yet highly desirable task. Recent advances introduce 2D diffusion priors, yielding reasonable results. However, existing methods are not hyper-realistic enough for post-generation usage, as users cannot view, render and edit the resulting 3D content from a full range. 
%
To address these challenges, we introduce \textbf{\ourmodel~} with several key designs and appealing properties:
\textbf{1) Viewable}: 360$^{\circ}$ mesh modeling with high-resolution textures enables the creation of visually compelling 3D models from a full range of observation points.
\textbf{2) Renderable:} Fine-grained semantic segmentation and data-driven priors are incorporated as guidance to learn reasonable albedo, roughness, and specular properties of the materials, enabling semantic-aware arbitrary material estimation.
\textbf{3) Editable: } For a generated model or their own data, users can interactively select any region via a few clicks and efficiently edit the texture with text-based guidance.
Extensive experiments demonstrate the effectiveness of \ourmodel~ in modeling region-aware materials with high-resolution textures and enabling user-friendly editing. 
We believe that \ourmodel~ holds promise for advancing 3D content creation and finding applications in various domains.

\end{abstract}

%
%


%
%


\begin{teaserfigure}
  \vspace{-0.15in}
  \centering
  \includegraphics[width=17.3cm]{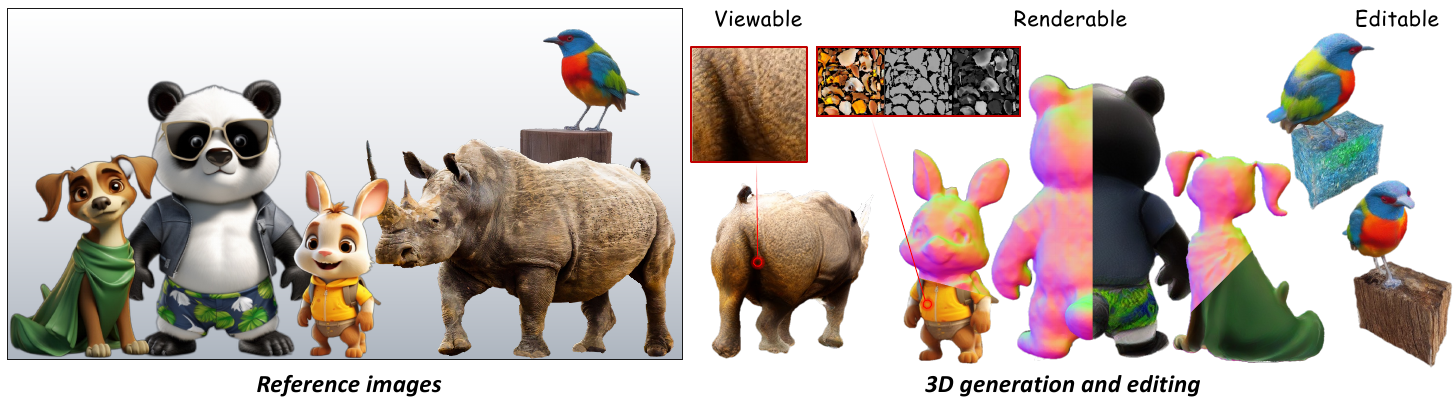}
  \vspace{-11pt}
  \caption{\textbf{Overview.} Given a single RGB image, we generate a realistic 3D model with rich details, which is full-range viewable, renderable, and editable.}
  \label{fig:teaser}
\end{teaserfigure}

\maketitle

\section{Introduction}
\label{sec:introduction}


In light of the high costs associated with expert-assisted 3D content creation and the increasing demand across diverse applications, such as gaming, online conferencing, and virtual social presence, there has been growing attention on 3D content generation, particularly in the domain of controllable generation. Traditional approaches~\cite{chan2021pi,deng2021deformed} in this field have predominantly relied on training category-specific models using large-scale 3D or 2D datasets, resulting in limited applications to specific categories. However, recent years have witnessed remarkable progress~\cite{poole2022dreamfusion,lin2023magic3d}, notably through the incorporation of diffusion priors derived from state-of-the-art 2D generative models. These advancements have facilitated the generation of reasonably accurate 3D content, marking a significant breakthrough in the field.


In recent 2D diffusion-based 3D content generation methods~\cite{poole2022dreamfusion,tang2023make}, it becomes common practice to incorporate text or single image conditions to achieve controllable generation. Due to its inherent ill-posed nature, researchers rely on a 2D diffusion model~\cite{rombach2021highresolution} as a guide prior to directing the rendering process, ensuring that all generated images are concentrated within the high-realism regions of the latent space. By confining the generated content to these regions, the overall realism of the produced 3D content is significantly enhanced.



Despite notable advancements, Current methods for 3D content generation suffer from two major drawbacks: limited post-generation usability and 2D diffusion bias. The former stems from the use of implicit 3D representations that trade off usability for fidelity. Users are unable to freely zoom, re-render, or edit the resulting 3D content to get the desired 3D content, which hampers its practical applicability and restricts creative possibilities. The latter arises from the training of the diffusion model on a 2D dataset that contains rich lighting and shading variations. These variations enhance the realism of the 2D images, but also introduce unwanted effects in the textures of the 3D models, as shown in Figure~\ref{fig:albedo_reg}-d. 



To address the above issues, we propose \ourmodel~, a 3D content generation and editing framework that is full-range viewable, renderable, and editable. 1) Full-range viewable: A novel custom super-resolution module is introduced, which incorporates pseudo multi-view images to facilitate high-resolution supervision. This module enables the generation of high-resolution textures for 360$^\circ$ content, allowing the creation of visually captivating 3D models from a full range of observation points. 2) Full-range renderable: The Segment-Anything-Model~\cite{kirillov2023segany} is integrated into our generation approach, enabling online 3D semantic segmentation. Leveraging the segmentation mask, we introduce a semantic-aware albedo regularization loss to mitigate the diffusion bias. To enable a more realistic rendering in downstream applications, we model the appearance using a spatially varying Bidirectional Reflectance Distribution Function (BRDF)~\cite{chen2022tango} and learn reasonable albedo, roughness, and specular properties of the materials,
enabling semantic-aware arbitrary material estimation. 3) Full-range editable: An interactive editing method is introduced, enabling users to perform interactive segmentation on 3D meshes effortlessly. By leveraging a normal-to-image model diffusion model, \ourmodel~ allows users to edit textures of specific regions in 3D meshes using text-based guidance. With just a few clicks, users can efficiently modify the targeted region, enhancing the editability and flexibility of the \ourmodel~.

Extensive experiments demonstrate the effectiveness of \ourmodel~ in modeling region-aware materials with high-resolution textures, and facilitating user-friendly editing, and show that \ourmodel~ surpasses state-of-the-art methods by a significant margin in terms of both 3D generation and editing quality. 
We believe that HyperDreamer, with its markedly superior quality and flexibility, effectively broadens the accessibility of AI-generated 3D content for practical applications.



\begin{figure*}[t]
	\centering
	\includegraphics[width=1.0\linewidth]{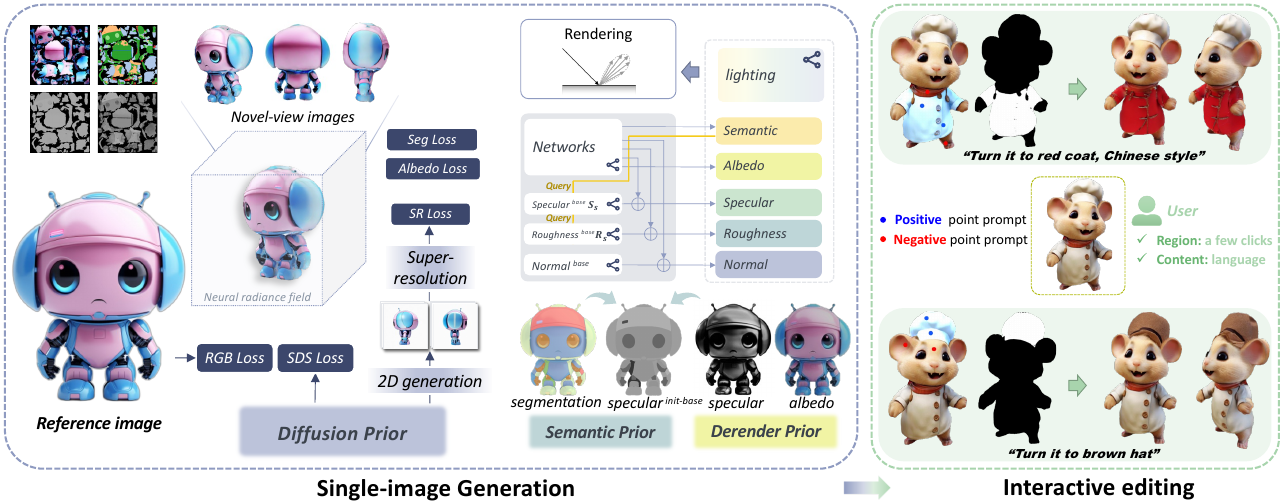}
        \setlength{\abovecaptionskip}{-3mm}
	\caption{\small
	\textbf{Overview of our 3D generation and editing pipeline.} We introduce diffusion priors, semantic priors, and derendering priors into this highly under-constraint problem to enable high-resolution textures with material modeling and interactive editing after the generation.}
	\label{fig:pipeline}
        \vspace{-2pt}
\end{figure*}




\section{Related Works}
\label{sec:related_works}

\paragraph{Text-guided 3D Generation.}
The text-guided 3D generation has gained significant attention following the remarkable success of text-to-image generation methods. 
Dream Fields~\cite{jain2021dreamfields} employed the text-image model CLIP~\cite{radford2021learning} to optimize NeRFs~\cite{mildenhall2020nerf} by aligning the text and image embeddings.
Building on the same principle, DreamFusion~\cite{poole2022dreamfusion} replaced CLIP with diffusion models and devised an SDS loss to distill knowledge from the denoising procedures.
Magic3D~\cite{lin2023magic3d} further enhanced generation performance by employing a coarse-to-fine framework and using meshes as the 3D representation in the second stage.
Fantasia3D~\cite{chen2023fantasia3d} disentangled the geometry and appearance modeling and introduced the spatially varying bidirectional reflectance distribution function (BRDF) for photo-realistic texture. Our approach utilizes a single image as the guided condition instead of text, which provides more detailed and specific information and introduces additional challenges.



%
\vspace{-5pt}
\paragraph{Single-image Reconstruction.}
%







 

Reconstructing 3D models from a single image has been a long-existing topic. 
Inference-based methods~\cite{choy20163d, tulsiani2017multi, melaskyriazi2023projection, gu2023nerfdiff, pavllo2023shape, vasudev2022ss3d, wu2023multiview, nichol2022pointe, jun2023shap-e} heavily depend on the datasets used for training, many of which can not handle diverse and general objects.
Optimization-based methods utilize priors from 2D text-to-image diffusion model to guide the reconstruction process.
RealFusion~\cite{melaskyriazi2023realfusion} employs textual inversion technique to bridge the gap between the reference image and text-conditioned guidance.
Make-it-3D~\cite{tang2023make} employs a two-stage framework and leverages high-quality textures extracted from the reference image.
Zero-1-to-3~\cite{liu2023zero123} synthesizes novel views by fine-tuning diffusion models~\cite{rombach2021highresolution} with multi-view data. 
It has also been applied to single-image 3D reconstruction by applying SJC~\cite{wang2022sjc}. Our work utilizes Zero-1-to-3 as the guidance model and incorporate several key designs to enable broader applications.

%
\vspace{-5pt}
\paragraph{Material and Illumination Estimation.}
Multi-view reconstruction methods~\cite{munkberg2022extracting} benefited from separately modeling geometry, material, and illumination conditions, while it's a highly ill-posed problem for generation. Previous works like Fantasia3D~\cite{chen2023fantasia3d} propose to learn globally varying roughness and metallic distributions, which may not always align with realistic material properties. Based on the material estimation approaches from a single image~\cite{wimbauer2022derendering, sang2020single}, we further propose a more plausible assumption that materials within the same semantic class share similar material properties, enabling spatially varying materials modelling while preventing degenerate solutions.


%
\vspace{-5pt}
\paragraph{Text-guided 3D Editing.}
%


Recently, text-guided image processing has experienced rapid development in both quality and diversity. 
Text2Mesh~\cite{michel2022text2mesh} proposes a neural style field, which uses CLIP to guide the initial mesh based on text. TANGO~\cite{chen2022tango} follows a similar scheme and uses a BRDF to optimize the appearance. 
However, there is a gap from the actual use due to insufficient accuracy.
More recently, TEXTure~\cite{richardson2023texture} leverages an improved depth-to-image diffusion process and applies an iterative scheme that paints a 3D model from different viewpoints. 
However, none of them enable text-guided editing of a local area on a 3D object. 
We propose an interactive editing method that users can edit textures based on text guidance in selected 3D regions with a few simple clicks or in a global manner.


%
\section{Preliminaries}
\label{sec:preliminaries}
\subsection{3D Representation}
Inspired by Magic3D~\cite{lin2023magic3d}, we adopt NeRF~\cite{mildenhall2020nerf} and DMTet~\cite{shen2021dmtet} for the first and second stage training, respectively. 
In the first stage, NeRF represents the scene as an implicit function that maps a 3D location $x$ and a 2D viewing direction $d$ to a volume density $\tau$ and color $c$. To render a pixel, NeRF alpha-composites the densities and colors along the ray that is cast from the camera to the pixel:
\begin{equation}
    C = \sum_k \alpha_k \prod_{k'<k}(1-\alpha_{k'}) c_k, \quad \alpha_k = 1 - \exp(-\tau_k \|x_{k+1} - x_k\|).
\end{equation}
To accelerate the training, we employ the efficient hash grid encoding from Instant NGP~\cite{muller2022instant} instead of pure MLPs. 

In the second stage, we adopt DMTet to produce high-resolution outputs without high computational and memory requirements. DMTet is a hybrid representation that integrates implicit and explicit surface representations and can efficiently render high-resolution textured meshes with differentiable rasterization. Formally, DMTet models the 3D shape as a deformable tetrahedral grid $(V_T, T)$, where $V_T$ are the vertices in the tetrahedral grid $T$. Each tetrahedron $T_k \in T$ has four vertices $\{v_{i_k}| i\in \{a,b,c,d\}\}$, each associated with a SDF value $s(v_i)$ and a deformation $\Delta v_{i_k}$. The surface mesh is extracted by differentiable marching tetrahedra algorithm. 
%
\subsection{Score Distillation Sampling (SDS)}
%
Previous works~\cite{poole2022dreamfusion, lin2023magic3d} have leveraged the 2D diffusion model~\cite{rombach2021highresolution} as prior knowledge for text-to-3D generation. The diffusion model $\phi$ learns a denoising function $\epsilon_{\phi}(x_t;y,t)$ that estimates the noise $\epsilon$ based on the noisy image $x_t$, text embedding $y$ and noise step $t$. It progressively reduces the noise and introduces image structure. To optimize the 3D scene $\theta$, Score Distillation Sampling (SDS) guides all rendered images to match the given text embedding $y$ under diffusion priors:
\begin{equation}
    \nabla_{\theta} \mathcal{L}_{SDS}(\phi, x=g(\theta)) = \mathbb{E}_{t,\epsilon}\left[\omega(t)(\epsilon_{\phi}(x_t;y,t) - \epsilon)\frac{\partial x}{\partial \theta}\right],
\end{equation}
where $g$ denotes the image renderer and $\omega(t)$ represents a weighting function. 
In addition to text conditional SDS, zero-1-to-3~\cite{liu2023zero123} introduces a 3D-aware SDS that conditions on input view and relative camera extrinsic to exploit the 3D consistent priors:
\begin{equation}
    \nabla_{\theta} \mathcal{L}_{SDS}(\phi, x=g(\theta)) = \mathbb{E}_{t,\epsilon}\left[\omega(t)(\epsilon_{\phi}(x_t;x_i,R,T,t) - \epsilon)\frac{\partial x}{\partial \theta}\right],
\end{equation}
where $x_i$ represents the input view, $R$ and $T$ are the relative camera rotation and translation from the input view to the desired viewpoint. 

%
\subsection{Segment Anything Model (SAM)}
%
Segment Anything Model (SAM)~\cite{kirillov2023segany} is the foundation model for general image segmentation, which supports various segmentation modes such as automatic everything and manual prompt. 
Taking point prompts as an example, SAM takes an image $I$ and a set of user-specific prompts $\mathcal{P}=(p,l)$ as inputs, and the output is a corresponding segmentation mask $M$. Among them, $\mathcal{P}$ includes $p$ and $l$, where $p$ is the set of each point coordinate and $l$ is the set corresponding to each point label. 
We use $S$ to represent the SAM model, so we have $M_{I,\mathcal{P}} = S(I, \mathcal{P})$. 

%
\section{Methodology}
\label{sec:method}
This section elaborates on the proposed framework in detail. Despite the inherent challenges posed by the ill-posed nature of the problem, HyperDreamer capitalizes on the deep priors from the 2D diffusion model, semantic segmentation model, and material estimation model, which collectively empower the capability for full-range viewing, rendering, and interactive editing. Specifically, (1) to achieve high-fidelity texture generation, we utilize high-resolution pseudo multi-view images for auxiliary supervision, as detailed in Sec~\ref{sec:SR}. 
(2) For material modeling, we introduce online 3D semantic segmentation and semantic-aware regularizations, which is initialized via material estimation results, as described in Sec.~\ref{sec:SAME}. 
(3) Furthermore, a novel interactive editing approach is proposed in Sec.~\ref{sec:edit} for effortless targeted modification of regions on 3D meshes via interactive segmentation.


\begin{figure}
	\centering
	\includegraphics[width=1.0\linewidth]{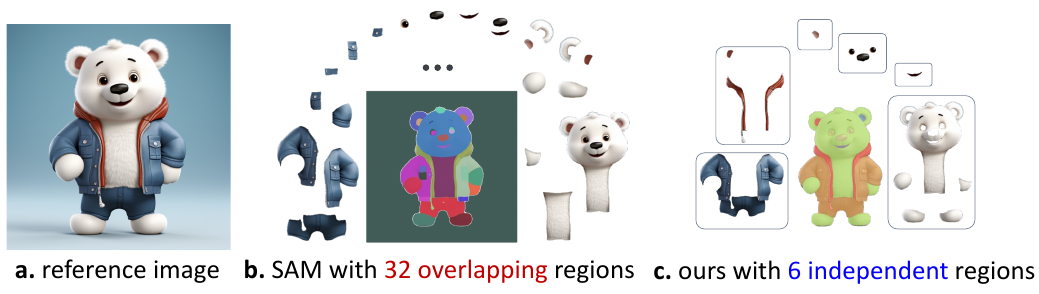}
        \setlength{\abovecaptionskip}{-3mm}
	\caption{\textbf{SAM at the generation stage.} We effectively cluster concise semantic groups compared to the raw SAM results.}
	\label{fig:sam}
        \vspace{-18pt}
\end{figure}

\subsection{360$^\circ$ High-Resolution Texture Generation} \label{sec:SR}
%
In the second training stage, the mesh representation allows for rapid images rendering, unlocking the potential of achieving high-resolution texture maps.
However, our guidance model, Zero-1-to-3~\cite{liu2023zero123}, was originally trained on low-resolution images ($256 \times 256$).
The resulting SDS loss fails to handle higher-resolution images, thereby limiting the benefits offered by mesh representation. The disparity between the resolutions used for training and inferencing leads to a relatively blurry texture map.

To overcome this challenge, we propose a high-resolution texture generation module. We first select a set of novel views and directly generate $m$ images per view using Zero-1-to-3. Subsequently, we employ a super-resolution network~\cite{rombach2021highresolution} to upscale the sampled images, enabling high-resolution supervision. Since the multi-view images generated by Zero-1-to-3 are not perfectly 3D consistent, directly applying per-pixel loss can lead to network instability.  Instead, we employ perceptual loss~\cite{johnson2016perceptual} in the feature space. By leveraging perceptual loss, we can minimize the content and style differences between two images without relying on pixel-level alignment, effectively alleviating inconsistencies during the training process.

\subsection{Semantic-Aware Material Estimation}
\label{sec:SAME}

\subsubsection{Online global semantic segmentation.}
During the second training stage, we also propose to integrate a new MLP-based branch upon the hash encoding and equip the framework with a globally-consistent mesh segmentation for further semantic regularization.
We first use SAM to produce over-segmented results of the reference image (Figure~\ref{fig:sam} (b)), and then we cluster different semantic parts by thresholding the feature similarity among them before assigning the semantic labels, as shown in Figure~\ref{fig:sam} (c).
We assume that the reference image already contains all of the semantic components of the generated 3D model. We also assign pseudo labels to novel view images by thresholding the feature similarities, and all these 2D labels are used to supervise the semantic branch training. We present detailed implementations in the supplementary materials.


\begin{figure}[t]
	\centering
	\includegraphics[width=1.0\linewidth]{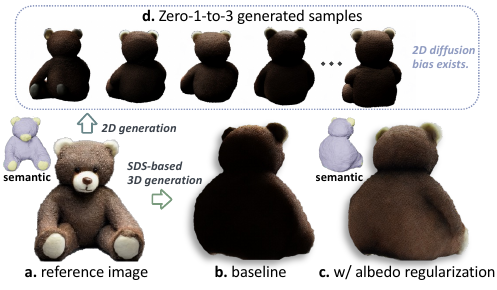}
        \setlength{\abovecaptionskip}{-3mm}
	\caption{\small
	\textbf{Diffusion bias.} The 2D diffusion bias in \textbf{d} leads to 3D generation failures in \textbf{b}, which can be alleviated by the albedo regularization in \textbf{c}.
	}
	\label{fig:albedo_reg}
    \vspace{-19pt}
\end{figure}

\subsubsection{Semantic-Aware Albedo Regularization.}
Recent approaches in the single-image 3D generation~\cite{melaskyriazi2023realfusion,tang2023make,liu2023zero123} optimize the model with diffusion priors and RGB reconstruction loss. They adopt different types of shading augmentations at novel views, including albedo, diffuse, and textureless, while only albedo shading is applied at the reference view. However, this pipeline introduces two inherent problems.

Firstly, the diffusion priors suffer from intrinsic shading and reflectance effects. For instance, Stable Diffusion and Zero-1-to-3 are trained on abundant images with lighting and shading variations, inevitably baking these effects into the textures of the generated 3D models. As shown in Figure~\ref{fig:albedo_reg}, given the front view of the teddy bear, Zero-1-to-3 tends to generate a dark back view, as if the light source only exists in the front, leading to a black back of the 3D model as in Figure~\ref{fig:albedo_reg}-b.

Secondly, the shading and reflectance characteristics in the reference image are integrated into the albedo color learning of the model via RGB reconstruction loss, making it challenging for re-rendering. 

We aim to introduce several albedo losses to alleviate the aforementioned problems. For the diffusion bias, we assume that albedo colors in regions under the same semantic label are similar. For the $N_s$ semantic labels, we maintain a albedo library called $A_s$, which is updated regularly according to the semantic-region-averaged albedo colors of the reference image along the training.
For each novel view, we predict segmentation masks with the semantic branch, and then we use a Gaussian filter to gain a weighted average of predicted albedo colors inside each semantic group. We propose a \textit{semantic-aware albedo regularization} as below:
\begin{equation}
    L_a = \sum\limits_{i=1}\limits^{N_s}||F_{gaussian}(A_{pred}^i) - A_s^i||_2^2.
\end{equation}

Furthermore, we incorporate a state-of-the-art single-image derendering framework~\cite{wimbauer2022derendering} to generate the albedo map of the reference image as an additional albedo supervision. 



\subsubsection{Appearance Modeling.}

To enable a more realistic rendering, we introduce the Physically-Based Rendering (PBR) material model. Following TANGO~\cite{chen2022tango} and PhySG~\cite{physg2021}, we leverage spatially varying BRDF (SVBRDF) to parameterize the material, including roughness, specular, and normal.
Based on the rendering equation~\cite{kajiya1986rendering}, given a location $x$ and the surface normal $n$, the incident light intensity at this point is denoted as $L_i(\omega_i;x)$ along the direction $\omega_i$; BRDF $f_r(\omega_o,\omega_i;x)$ denotes the reflectance coefficient of the material viewing from direction $\omega_o$. The observed light intensity $L_o(\omega_0;x)$ is calculated over the hemisphere $\Omega=\{\omega_i:\omega_i \cdot n > 0\}$:
\begin{equation}
    L_o(\omega_0;x) = \int_\Omega L_i(\omega_i;x) f_r(\omega_o,\omega_i;x) (\omega_i \cdot n) d\omega_i .
    \label{eq:render_equation}
\end{equation}

We utilize spherical Gaussians (SGs)~\cite{yan2012accurate} to approximate the rendering equation in closed form. For a spherical Gaussian with $n$ dimensions, given the lobe axis $\xi \in \mathbbm{S}^2$, lobe sharpness $\lambda \in \mathbbm{R}_+$, and lobe amplitude $\mu \in \mathbbm{R}^n_+$, the spherical function is formulated as:
\begin{equation} 
    G(\nu;\xi,\lambda,\mu) = \mu e^{\lambda (\nu \cdot \xi - 1)},
    \label{eq:spherical_function}
\end{equation}
where $\nu \in \mathbbm{S}^2$ denotes the input.

The environment map $L(\omega_i)$ is represented as a mixture of SGs:
\begin{equation}
    L_i(\omega_i) = \sum\limits_{k=1}^{M} G(\omega_i;\xi_k,\lambda_k,\mu_k).
    \label{eq:environment_map}
\end{equation}

The SVBRDF is divided into diffuse BRDF and specular BRDF: $f_r(\omega_o,\omega_i;x) = f_d(x)/\pi + f_s(\omega_o,\omega_i;x)$. The diffuse term is modeled as an MLP based on the multi-resolution hash input encoding. 
And the specular term at location $x$ is formulated as:
\begin{equation}
    f_s(\omega_o,\omega_i;x) =  G(h;n_x,\frac{\lambda_x}{4h \cdot \omega_o},\mathcal{M}_x\mu_x),
    \label{eq:specular_tern}
\end{equation}
where $h$ is half vector and $\mathcal{M}$ is Fresnel and shadowing effects.

The last term in Eqn.~\ref{eq:render_equation} is approximated as~\cite{meder2018hemispherical}: $(\omega_i \cdot n) = G(\omega_i;0.0315,n,32.7080) - 31.7003$.

Therefore, the rendering equation is represented as the multiplication of SGs and can be calculated in closed form. Learnable parameters above include $\{\xi_k,\lambda_k,\mu_k\}_{k=1}^M$ for the environmental map, diffuse albedo $f_d$, and the spatially varying $\{\lambda, \mu\}$. We assume that regions with the same semantic label usually share alike materials and enforce channel consistency in roughness and specular. Please refer to the supplementary materials for more details.

\begin{figure}
	\centering
	\includegraphics[width=1.0\linewidth]{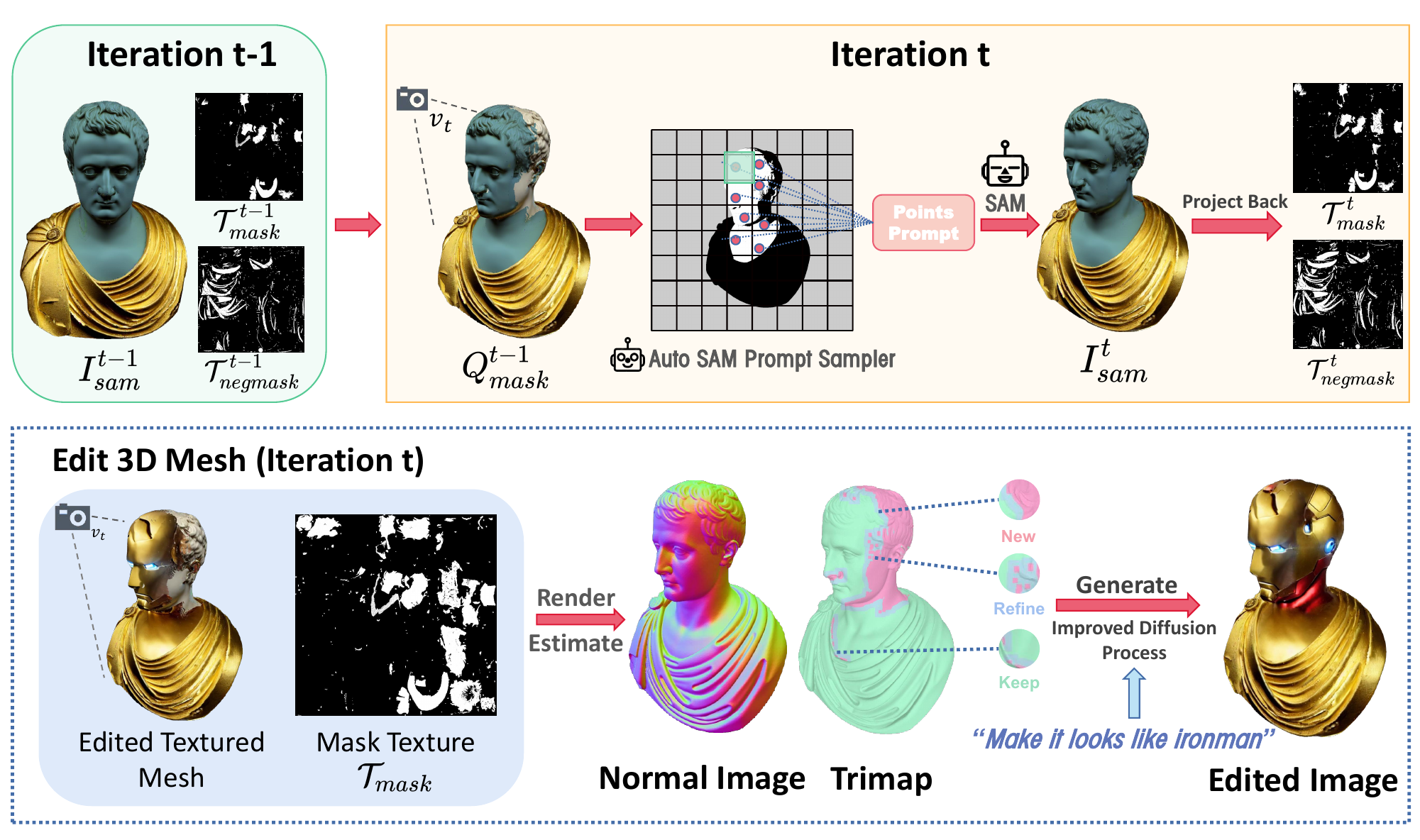}
        \setlength{\abovecaptionskip}{-3mm}
	\caption{\small
	\textbf{Interactive editing process.} Users can select the interest regions and then our method output the texture mask of the target area to our texture synthesis pipeline for text-guided editing.
	}
	\label{fig:editing}
    \vspace{-15pt}
\end{figure}

\begin{figure*}[t]
	\centering
	\includegraphics[width=1.0\linewidth]{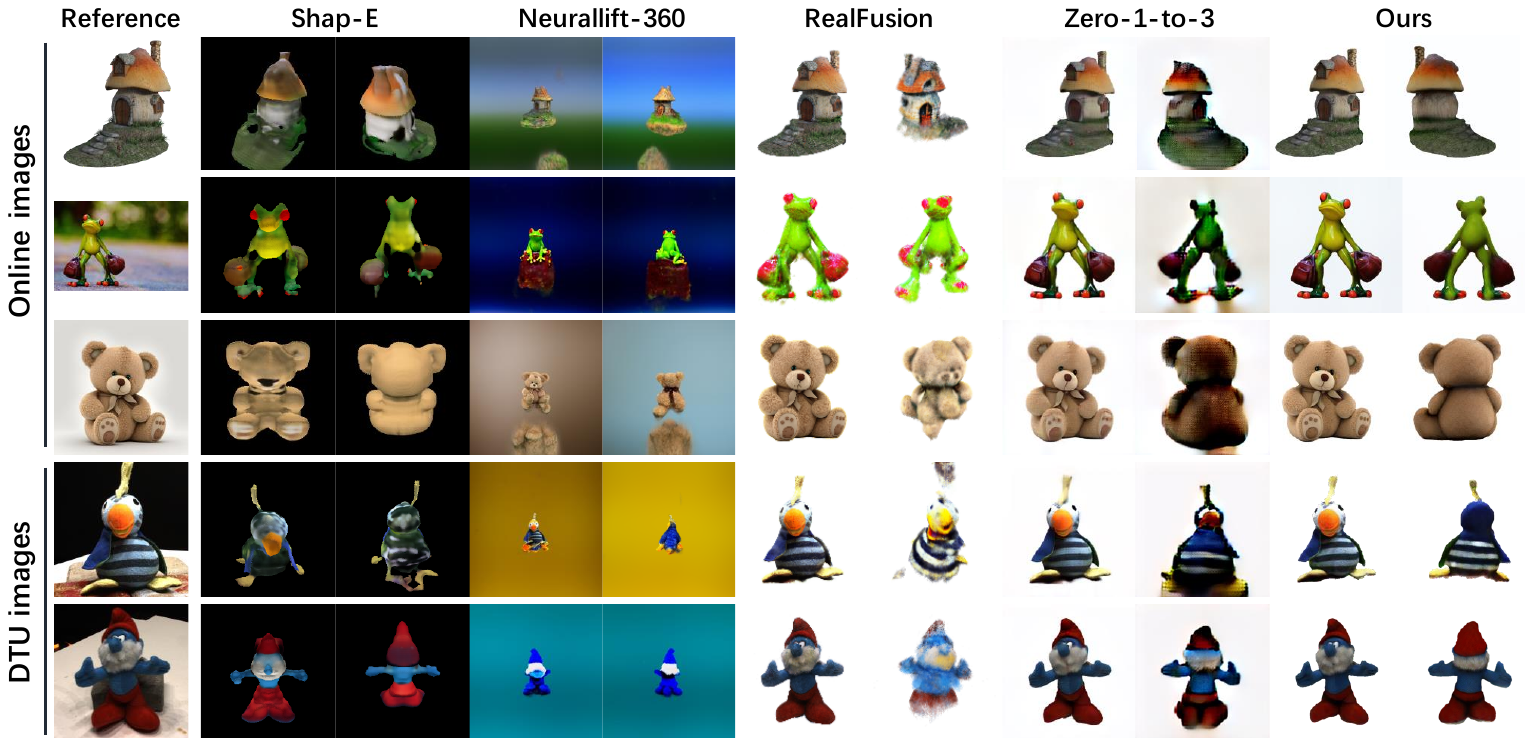}
        \setlength{\abovecaptionskip}{-3mm}
	\caption{\small
	\textbf{Qualitative comparisons.} \ourmodel~ generates a high-fidelity reference view and more realistic and reasonable results at the novel view.
	}
	\label{fig:exp_main}
    \vspace{-10pt}
\end{figure*}

\subsection{Interactive Editing}
\label{sec:edit}
Editing a 3D model requires complex interaction with 3D shapes while maintaining global consistency to achieve the desired design. We propose an intelligent and user-friendly interactive 3D editing tool that allows users to quickly settle the target area in 3D space by one-shot selection and edit its texture based on text guidance.

\subsubsection{Interactive Segmentation In Mesh}

Interactive segmentation in the 3D mesh enables users to segment any region of the 3D object. In our method, as shown in Figure~\ref{fig:editing}, we use two UV maps to represent the masks of the 3D mesh, where $\mathcal{T}_{mask}$ for selected regions and $\mathcal{T}_{negmask}$ for remaining regions, respectively. Given a target view $v_t$, we can render the masks $Q^{t-1}_{mask}$ and $Q^{t-1}_{negmask}$ which are actually the point prompts cache from the previous $t$ - 1 views but not complete segmentation results in the current view. Then, we sample points with a patch sampling mechanism on $Q^{t-1}_{mask}$ as positive prompts and $Q^{t-1}_{negmask}$ as negative prompts in each patch to generate refined segmentation results $I_{sam}$ and $I_{negsam}$ via SAM. Inverse rendering is then applied to project $I_{sam}$ and $I_{negsam}$ onto ${\mathcal{T}}_{mask}$ and ${\mathcal{T}}_{negmask}$, where we use a gradient-based optimization to $\mathcal{T}_{mask}$for $\mathcal{L}_t$ over the values of $\mathcal{T}_{mask}$ when rendered through the differential renderer ${\mathcal{R}}$ ~\cite{jatavallabhula2019kaolin}.That is, 

\begin{equation}
\nabla_{{\mathcal{T}}_{mask}^{t}} \mathcal{L}_{t}=\left[\left(\mathcal{R}\left(\text { Mesh }, {\mathcal{T}}_{mask}^{t}, v_{t}\right)-I_{sam}^{t}\right) \odot m_{t}\right] \frac{\partial \mathcal{R} \odot m_{t}}{\partial {\mathcal{T}}_{mask}^{t}},
\end{equation}
where, $m_t$ is the mask of mesh at the view $v_t$. Similarly, the method of projecting back to texture in each view in subsequent is the same.

\subsubsection{Text-Guided Texture Synthesis}
We apply Normal-to-Image model $\mathcal{M}_{normal}$ based on ControlNet~\cite{zhang2023adding} to paint textures that closely match the surface details on the 3D Mesh directly.

To address the inconsistency problem,
we divide each rendered view into three partitions: $M_{new}$, $M_{keep}$, and $M_{refine}$. The $M_{new}$ partition is the target region that needs to be painted for the first time. The $M_{keep}$ partition is either a previously well-painted target region or a region that is out of the target region. The $M_{refine}$ partition is the region painted from the previous views, but they are mainly the junction of adjacent views and need further refinement.



\begin{table}[t]
  \centering
  \caption{Quantitative results on our data.}
    \vskip -0.1in
    \small
    \begin{tabular}{l|ccc}
    \toprule
    Method  & Contextual $\downarrow$ & CLIP $\uparrow$ & Perceptual $\downarrow$ \\
    \midrule
    Shap-E & 4.95  & 0.68  &  -  \\
    NeuralLift-360 & 4.71  & 0.78  & 0.67 \\
    RealFusion & 2.25  & 0.79  & 0.17 \\
    Zero-1-to-3 & 3.36  & 0.74  & 0.13 \\
    \midrule
    \textbf{Ours} & \textbf{2.11} & \textbf{0.86} & \textbf{0.10} \\
    \bottomrule
    \end{tabular}%
  \label{tab:custom_results}%
  \vspace{-10pt}
\end{table}%

To attain $M_{refine}$, we first perform an opening operation, $Open$, on the mask $M_{new}$ to eliminate out-lie small regions. We then performs erode $\mathcal{E}$ and dilate $\mathcal{D}$ as follows,
\begin{equation}
M_{refine} =\mathcal{D} (Open (M_{new})) - \mathcal{E} (Open (M_{new}))).
\end{equation}

In the ${\mathcal{M}}_{normal}$, we modify the sampling process by blender diffusion to inject the information of region partition into the denoising process. The mask $M_{paint}$ explicitly blends the noised latent $z_{Q_{t}}$ and the denoised latent estimation $z_t$ as follows:
\begin{equation}
M_{\text {paint}} = 
    \begin{cases}
        0 & M_{keep} \\
        1 & M_{new} \cup M_{refine},
    \end{cases}
\end{equation}
\begin{equation}
\hat{z}_{t}=\hat{z}_{t} \odot M_{paint}+z_{t} \odot(1-M_{paint}).
\end{equation}

Based on the above texture synthesis method, we can achieve local editing in the 3D mesh. In more detail, we can limit the editing area to the target region by doing the dot product with the original texture map $\mathcal{T}$ and the $\mathcal{T}_{mask}$ obtained by interactive segmentation in mesh. Finally, users can select any region in the 3D object to edit based on text guidance, as illustrated in Figure~\ref{fig:editing}.


\begin{table}[t]
  \centering
  \caption{Quantitative results on the DTU dataset.}
  \vskip -0.1in
  \small
    \begin{tabular}{l|ccc}
    \toprule
    Methods  & Contextual $\downarrow$ & CLIP $\uparrow$ & Perceptual $\downarrow$ \\
    \midrule
    Shap-E & 2.82  & 0.80  &  -  \\   
    NeuralLift-360 & 4.74  & 0.78  & 0.72 \\
    RealFusion & 4.84  & 0.82  & 0.47 \\
    Zero-1-to-3 & 4.50  & 0.80  & 0.43 \\
    \midrule
    \textbf{Ours} & \textbf{2.08} & \textbf{0.89} & \textbf{0.13} \\
    \bottomrule
    \end{tabular}%
  \label{tab:dtu_results}%
  \vspace{-15pt}
\end{table}%

%
\subsection{Implementation Details}
\subsubsection{Model and training.}
For the generation process, we follow Instant-NGP~\cite{muller2022instant} to adopt a two-stage training pipeline: a coarse NeRF is trained for 50 epochs in the first stage, guided by Zero-1-to-3 based SDS loss and other regularisation terms like depth and normal loss. We use MiDaS~\cite{Ranftl2021midas} and OmniData~\cite{eftekhar2021omnidata} to extract the depth and normal estimations, respectively. We train a DMTet~\cite{shen2021dmtet} for 100 epochs in the second stage based on the first stage model. The SR, semantics, and material modules are integrated with the second stage only. More details are presented in the supplementary materials.

\subsubsection{Methods for comparison.}
We compare four recent approaches for single-image based 3D generation for arbitrary images: 
\textbf{Shap-E}~\cite{jun2023shap-e} is a conditional generative model for implicit representations trained on millions of 3D assets;
\textbf{NeuralLift-360}~\cite{Xu_2022_neuralLift} generates the neural radiance field based on the CLIP-guided diffusion priors.
\textbf{RealFusion}~\cite{melaskyriazi2023realfusion} applies the SDS loss based on Stable Diffusion~\cite{rombach2021highresolution} and RGB reconstruction loss with the reference view, 
in comparison, \textbf{Zero-1-to-3}~\cite{liu2023zero123} leverages the SJC loss based on its viewpoint-conditioned model as the guidance.

\section{Experiments}
\label{sec:experiments}

\begin{figure}
	\centering
	\includegraphics[width=0.95\linewidth]{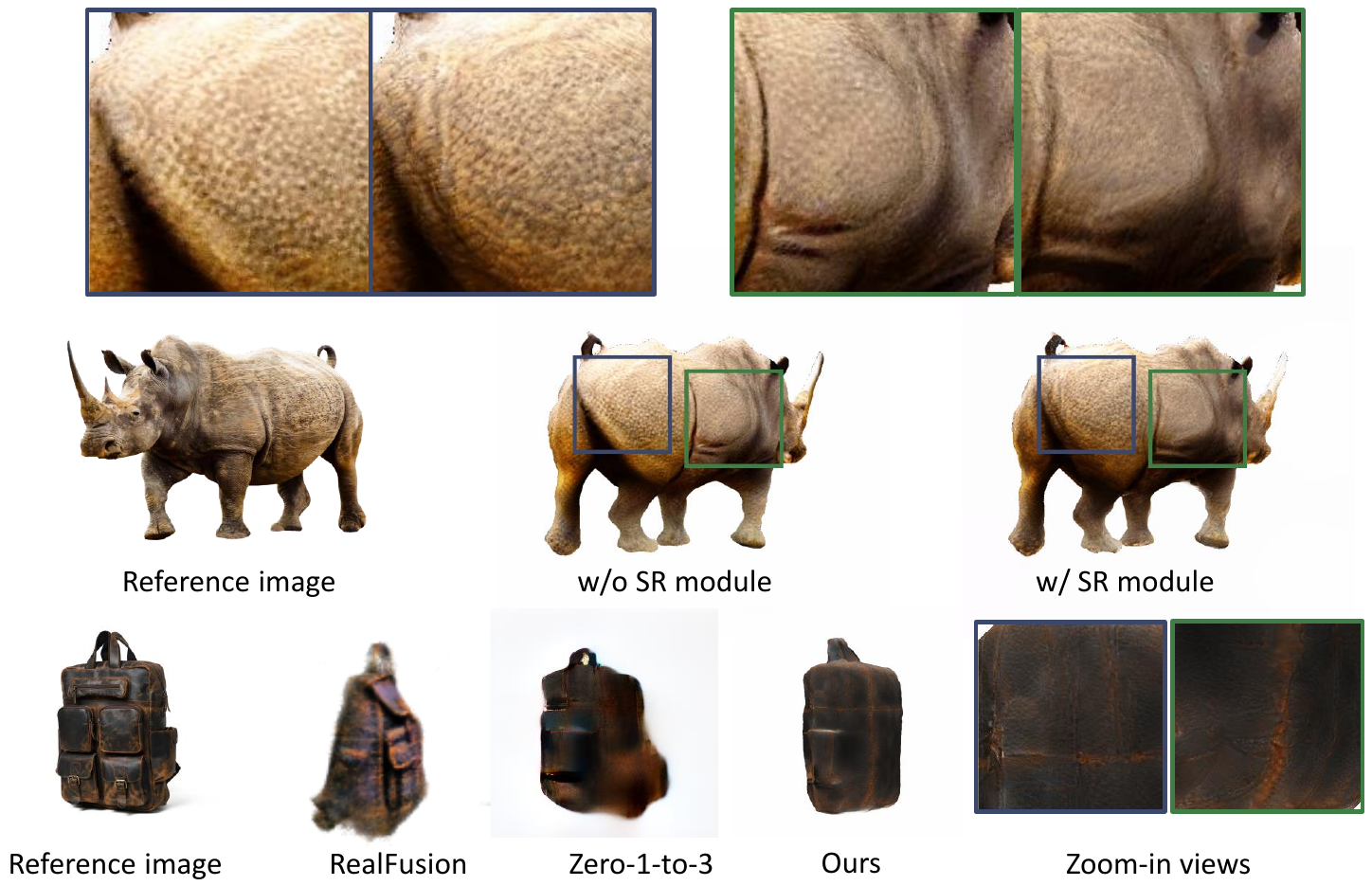}
        \setlength{\abovecaptionskip}{-0.1mm}
	\caption{\small
 \textbf{Ablation on the super-resolution (SR) module.} High-frequency details on textures are generated under SR supervision.
	}
	\label{fig:sr_module}
        \vspace{-5pt}
\end{figure}

\begin{figure}
	\centering
	\includegraphics[width=0.95\linewidth]{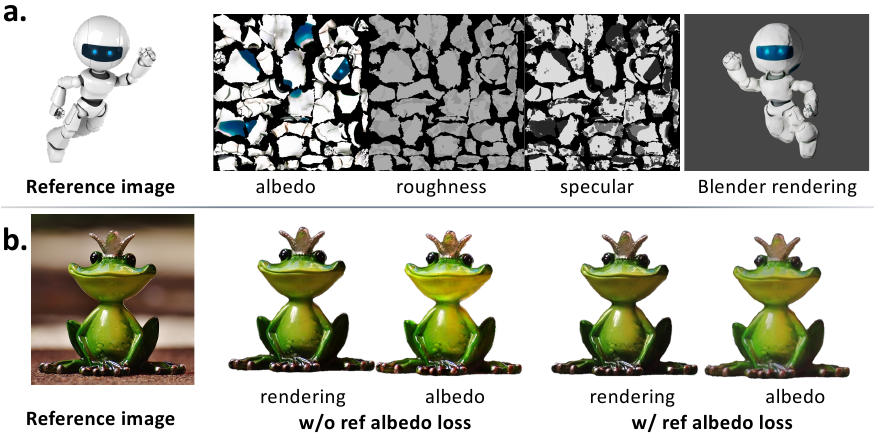}
        \setlength{\abovecaptionskip}{-0.1mm}
	\caption{\small
	\textbf{Analysis of the material modeling.} We show an example of the output roughness and specular maps in \textbf{a}, together with its rendering results in Blender. We show how the albedo loss at the reference view help alleviate shading and reflectance learning in the albedo texture in \textbf{b}.
	}
	\label{fig:exp_materials}
        \vspace{-10pt}
\end{figure}

\subsection{Qualitative Comparisons}
We show some qualitative comparisons with several state-of-the-art works in Figure~\ref{fig:exp_main}, where we present both the reference view and the back view of the object for each method. The results by Shap-E are relatively worse than the other optimization-based methods. The instances generated by NeuralLift-360 are small in size and low in quality, while the basic semantics is reserved. RealFusion and Zero-1-to-3, both leverage the reference view RGB reconstruction loss for constraint and thus keep a high fidelity with the reference image. While RealFusion suffers severely from the multi-face problem, and the results from Zero-1-to-3 are blurry. Our method achieves the highest quality in both the reference view and the back view, presenting realistic and reasonable generations.

\subsection{Quantitative Comparisons}
We adopt three metrics for quantitative comparisons: 1) LPIPS~\cite{johnson2016perceptual} evaluates the reconstruction quality of the reference view image; 2) Contextual distance~\cite{mechrez2018contextual} evaluates the pixel-level distance between the rendered novel view images and the reference image; 3) CLIP-Score~\cite{radford2021learning} measures the semantic-level distance between the novel view images and the reference image. We select 20 images online with a wide range of diversity, and we select 10 instances from the DTU dataset ~\cite{aanaes2016large} that are basically complete. The results for the two datasets are shown in Table~\ref{tab:custom_results} and Table~\ref{tab:dtu_results}, respectively. Our model outperforms the comparison methods in all three metrics by a large margin, quantitatively revealing the effectiveness of the pipeline.

\subsection{Analysis and Ablations}
\subsubsection{Super Resolution}
We show how our SR module works in Figure~\ref{fig:sr_module}. It largely enhances the texture details and realism in our model, which enables it to support high-resolution zoom-in views in comparison with other methods.

\vspace{-3pt}
\subsubsection{Materials}
Examples of the generated roughness and specular map are shown in Figure~\ref{fig:exp_materials}-a, where we observe that the material proprieties are highly correlated with the semantic label of the region. We also show how the albedo loss helps decompose the albedo texture out of the reference view.

%
\vspace{-3pt}
\subsubsection{Editing}

\begin{figure}
	\centering
	\includegraphics[width=0.95\linewidth]{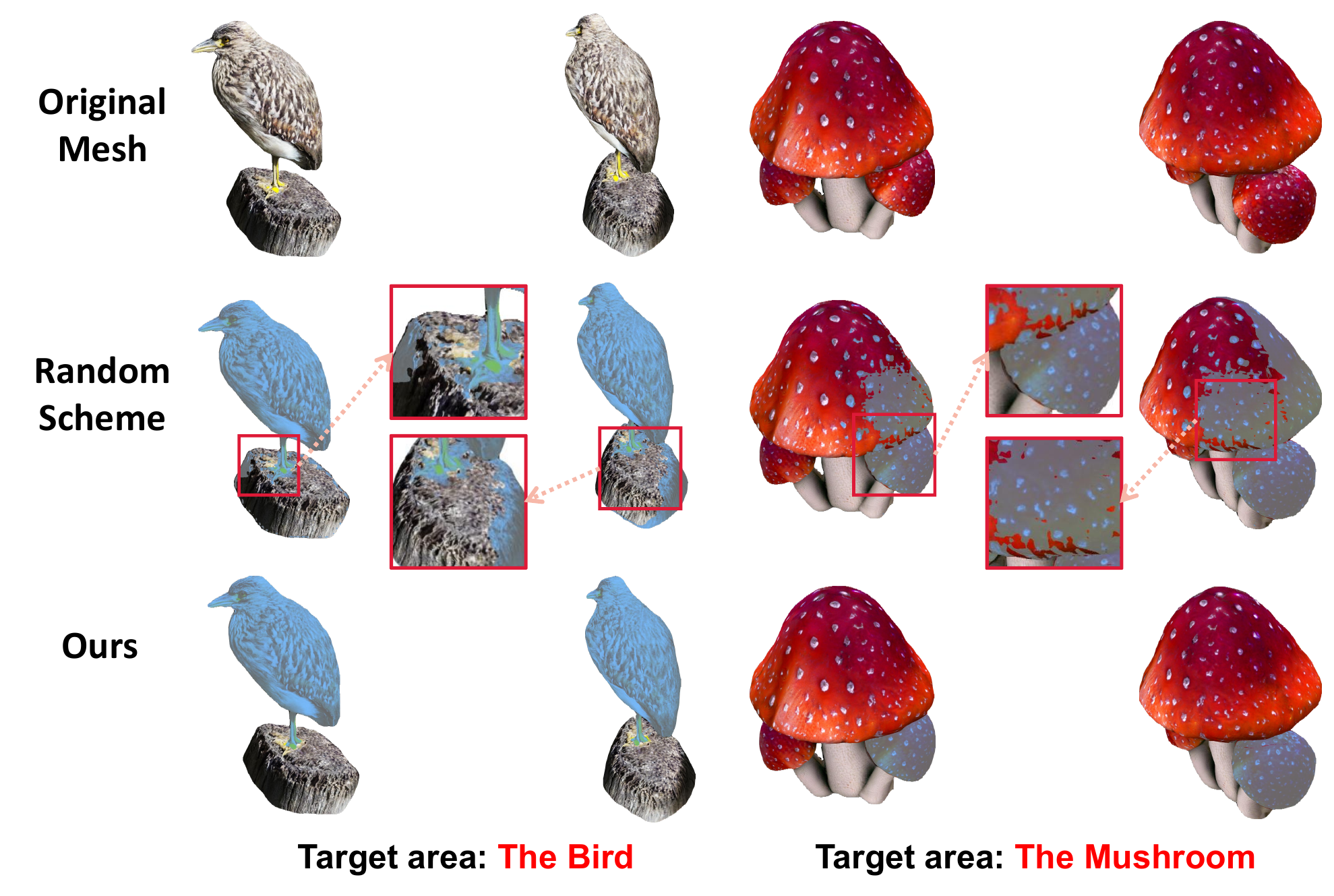}
        \setlength{\abovecaptionskip}{-0.1mm}
	\caption{\small
    \textbf{Analysis of our scheme of segmentation in mesh.} Our method has better ability to handle complex circumstances.
	}
	\label{fig:prompt_ablation}
        \vspace{-18pt}
\end{figure}

We show in Figure~\ref{fig:prompt_ablation} that the Naive method for segmentation in the mesh, which only inputs positive prompt and randomly samples the point prompts cache, has a high probability of failure especially in the condition of dealing with discrete and complex regions. However, adopting our scheme, it is more robust with patch sampling in the positive and negative prompts and input both into SAM.  
\section{Conclusion}
\label{sec:conclusion}
This paper introduces a framework, \textbf{\ourmodel~}, which enables hyper-realistic 3d content generation and editing for a single image. In contrast to previous works, the 3D content generated by our method is full-range viewable, renderable, and editable. 
Extensive experiments demonstrate the effectiveness of \ourmodel~ in modeling region-aware materials with high-resolution textures and enabling user-friendly editing. We believe that \ourmodel~ holds promise for advancing 3D content creation and editing, which would be practical for both academic and industrial usage.

\noindent{\textbf{Acknowledgement.}}
This project is funded by Shanghai AI Laboratory and the Ministry of Education, Singapore, under its MOE AcRF Tier 2 (MOE-T2EP20221- 0012), NTU NAP, and under the RIE2020 Industry Alignment Fund – Industry Collaboration Projects (IAF-ICP) Funding Initiative, as well as cash and in-kind contribution from the industry partner(s).
\clearpage


\bibliographystyle{ACM-Reference-Format}
\bibliography{egbib}


\begin{figure*}
    \centering
    \includegraphics[width=0.92\linewidth]{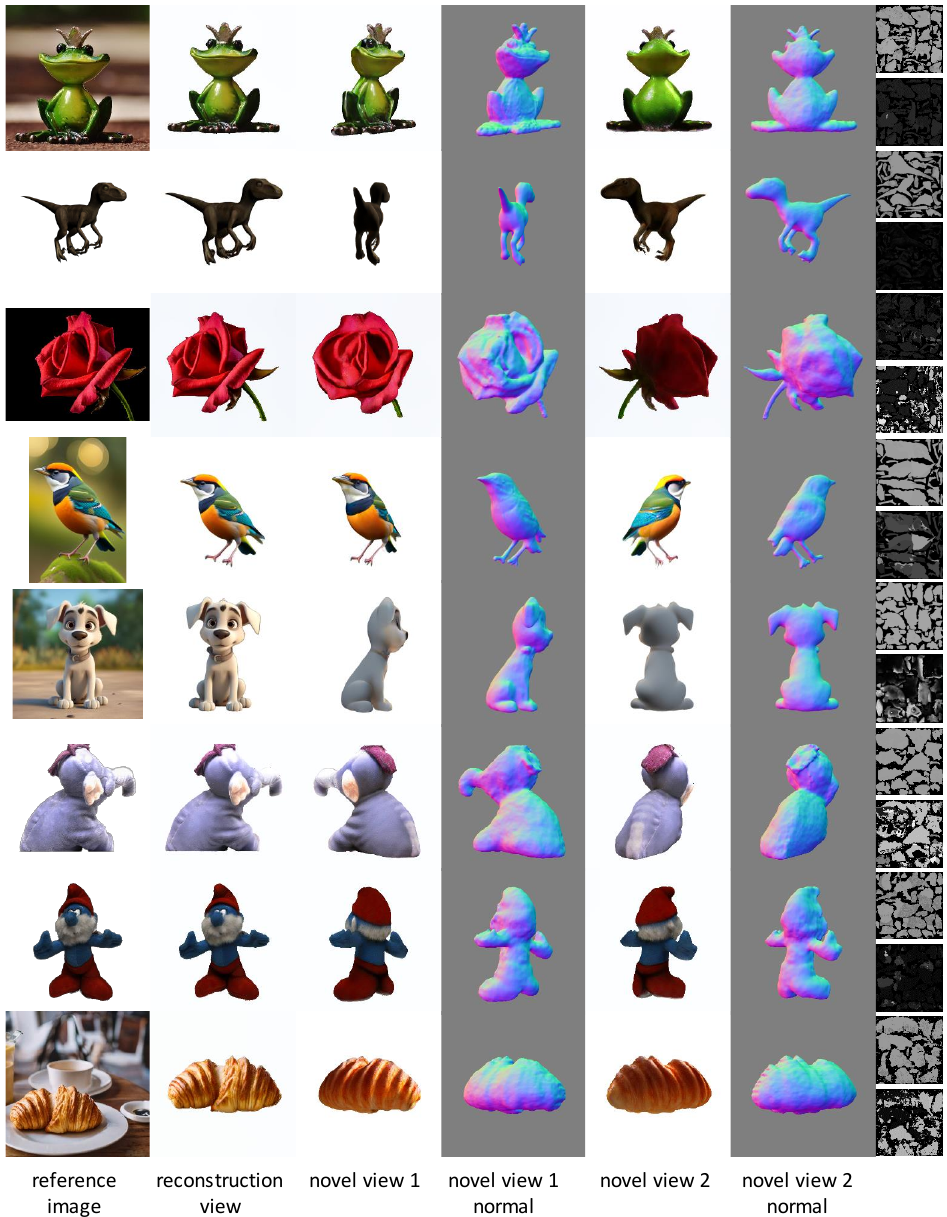}
    \caption{Additional results by \ourmodel~ with more views. Images in the  last column are specular and roughness map respectively (from top to bottom).}
    \label{fig:fig_only}
\end{figure*}

\end{document}